\documentclass{article}

\usepackage[preprint,nonatbib]{neurips2019}

\usepackage[utf8]{inputenc} 
\usepackage[T1]{fontenc}    
\usepackage{hyperref}       
\usepackage{cleveref}
\usepackage{url}            
\usepackage{booktabs}       
\usepackage{amsfonts}       
\usepackage{nicefrac}       
\usepackage{microtype}      

\usepackage{xcolor}
\usepackage{graphicx}
\usepackage{multirow}

\title{Panoptic Edge Detection}

\author{
Yuan Hu\thanks{ indicates equal contribution} \\
$^1$Institute of Remote Sensing and Digital Earth, CAS, Beijing 100094, China \\
$^2$University of Chinese Academy of Sciences, Beijing 100049, China \\
\texttt{huyuan@radi.ac.cn} \\
\And
Yingtian Zou$^*$ \\
National University of Singapore \\
\texttt{elezouy@nus.edu.sg}
\And
Jiashi Feng \\
National University of Singapore \\
\texttt{elefjia@nus.edu.sg}
}

\begin{document}

\maketitle

\vspace{-5mm}
\begin{abstract}
\vspace{-3mm}
 Pursuing more complete and coherent scene understanding towards realistic vision applications drives edge detection from category-agnostic to category-aware semantic level. However, finer delineation of instance-level boundaries still remains unexcavated. In this work, we address a new finer-grained task, termed panoptic edge detection (PED), which aims at predicting semantic-level boundaries for \emph{stuff} categories and instance-level boundaries for \emph{instance} categories, in order to provide more comprehensive and unified scene understanding from the perspective of edges.
We then propose a versatile framework, Panoptic Edge Network (PEN), which aggregates different tasks of object detection, semantic and instance edge detection into a single holistic network with multiple branches. Based on the same feature representation, the semantic edge branch produces semantic-level boundaries for all categories and the object detection branch generates instance proposals. Conditioned on the prior information from these two branches, the instance edge branch aims at instantiating edge predictions for \emph{instance} categories. Besides, we also devise a Panoptic Dual F-measure ($F^2$) metric for the new PED task to uniformly measure edge prediction quality for both \emph{stuff} and \emph{instances}. By joint end-to-end training, the proposed PEN framework outperforms all competitive baselines on Cityscapes and ADE20K datasets.

\end{abstract}

\vspace{-4mm}
\section{Introduction}
\vspace{-3mm}
Category-aware Semantic Edge Detection (SED) is now receiving increasing attention in computer vision with the growing demand for finer scene understanding systems in autonomous driving, robots, etc. This task predicts semantic-level boundaries for both \emph{stuff} categories (such as sky, building) and \emph{instance} categories (such as person, car) without distinguishing different instances.
But in real scenarios, instance-level boundaries of instance categories are often intensively desirable. This leads to a  finer-grained task, which we call panoptic edge detection (PED).

Panoptic edge detection can be considered as a non-trivial extension of semantic edge detection in producing instance-level boundaries for instance categories.
The commonality of both tasks lies in that for stuff categories, they produce only semantic-level boundaries. PED is by nature a multi-label problem, where each pixel can be associated with more than one stuff categories and/or more than one instances. For example, an edge pixel can also be associated with a stuff category and an instance, or two instances, or two stuff categories etc.

To tackle this new task, we propose a multi-branch framework named Panoptic Edge Network (PEN), which jointly optimizes the performance on semantic edge detection, object detection and instance edge detection. The corresponding three task branches share a basic visual representation and infer different forms of predictions based on different optimization objectives. In the semantic edge detection branch, PEN aggregates the hierarchical features to predict category-specific edges for both \emph{stuff} and \emph{instance} categories. For the object detection branch, PEN adopts a region-based object detection method to regress bounding boxes of instances following Faster R-CNN~\cite{ren2015faster}. Guided by semantic edges and bounding boxes, the instance edge detection branch will output the instance-specific edge. By assembling the above three branches into a single unified network, PEN is able to predict semantic-level boundaries for stuff categories as well as instance-level boundaries for instance categories at the same time.

We also devise a new evaluation metric for PED task called \emph{panoptic dual F-measure} ($F^2$), which uniformly measures the quality of semantic-level boundaries for stuff and instance-level boundaries for instances. For stuff, $F^2$ is equivalent to Maximum F-Measure (MF) at optimal dataset scale (ODS) which measures the quality of semantic-level boundaries in SED task. Regarding instances, $F^2$ is the combination of MF (ODS) of edge detection and $F_1$ score of object detection, which makes it qualified for measuring the  quality of instance-level boundaries. By setting $F_1$ to $1$ for stuff, we can obtain a unified evaluation protocol for both stuff and instance categories.

Our contributions can be summarized into three folds:
\begin{itemize}
    \item We address a new task named Panoptic Edge Detection (PED), which encompasses both semantic edge detection and instance edge detection. It is a more complete and finer-grained task setting in the field of edge detection.
    \item We propose a Panoptic Edge Network to solve PED. It integrates three branches, semantic edge detection branch, object detection branch and instance edge detection branch, into a single unified network to accomplish semantic-level boundary detection for stuff and instance-level boundary detection for instances at the same time.
    \item We propose a new panoptic edge metric named \emph{panoptic dual F-measure}($F^2$), which uniformly measures the quality of semantic-level boundaries for stuff and instance-level boundaries for instances.
\end{itemize}

\section{Related work}

Edge detection has evolved from category-agnostic to category-aware semantic level towards more complete and coherent scene understanding. Category-agnostic edge detection aims to detect object boundaries in a simple binary classification manner, in which each pixel is classified as edge or non-edge without distinguishing specific categories. Recently, deep methods~\cite{deng2018learning,xie2015holistically,liu2017richer,liu2016learning} employ holistically nested topology to solve this task such as HED~\cite{xie2015holistically}. Category-aware semantic edge detection~\cite{prasad2006learning} is an extension of category-agnostic edge detection. Inspired by HED, CASENet \cite{yu2017casenet} shares the same set of low-level hierarchical features and then fuses them with semantic channels at the top convolution layer. Several following works~\cite{liu2018semantic,yu2018simultaneous,hu2019dynamic,acuna2019devil} continuously lift semantic edge detection performance. In existing SED setting, predicted boundaries of \emph{instance} categories are semantic-level and in this work, we propose a new panoptic edge detection task aiming to form a complete and coherent scene understanding by predicting instance-level boundaries for \emph{instances}. We also design a multi-branch network PEN to solve this challenging task.


Our work is also related to panoptic segmentation~\cite{kirillov2018panoptic}, which unifies semantic segmentation and instance segmentation. Panoptic segmentation also generates a coherent scene parsing but from the perspective of segmentation. Panoptic FPN~\cite{kirillov2019panoptic} tackles panoptic segmentation by extending Mask R-CNN\cite{he2017mask} with a semantic segmentation branch. We choose Panoptic FPN with some minor modifications (denoted as \emph{Panoptic FPN*}) as a baseline model. The comparison experiments between \emph{Panoptic FPN*} with the proposed PEN model (in Section \ref{section4.4: Comparison with  State-of-the-Arts}) demonstrate that directly utilising a framework designed for panoptic segmentation will not work for panoptic edge detection due to the intrinsic differences between these two tasks.

\begin{figure*}[t]
\centering
\resizebox{\textwidth}{!}{
\includegraphics[]{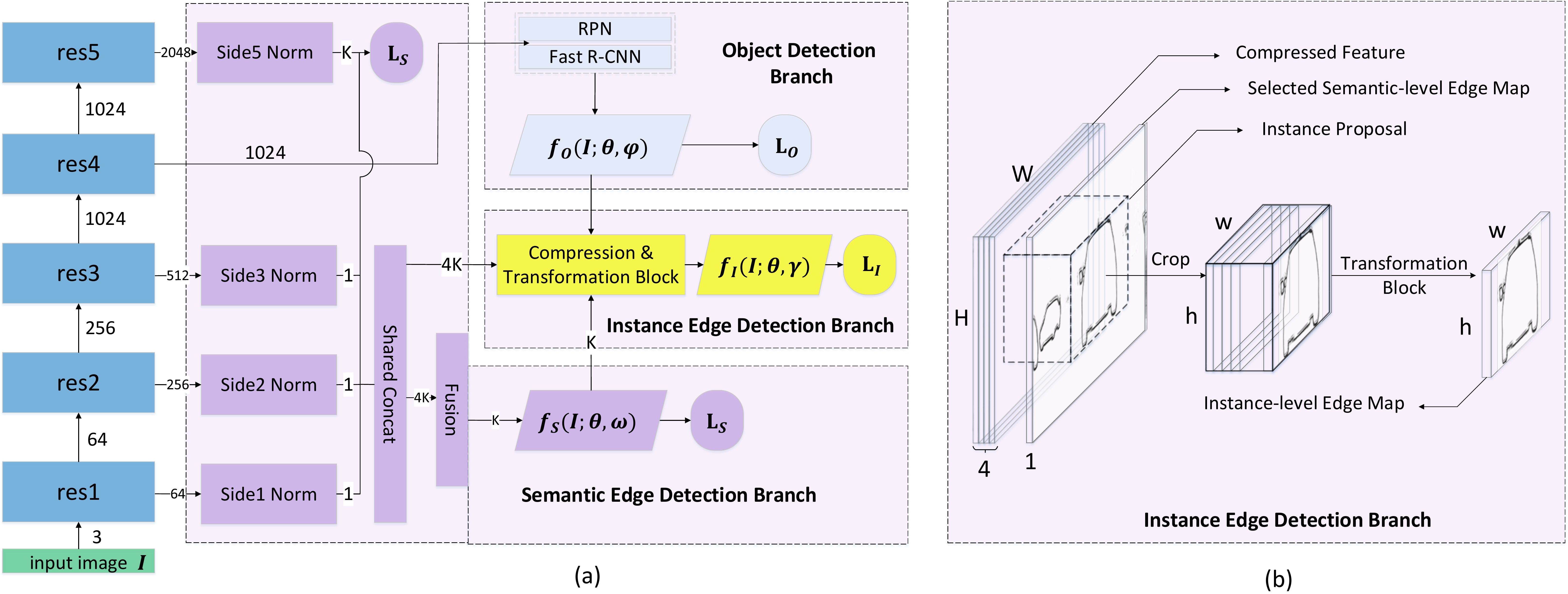}
}
\caption{Overall architecture of proposed PEN. (a) With ResNet-50 \cite{he2016deep} as backbone (blue blocks), three branches are introduced: semantic edge detection branch (purple blocks), object detection branch (cyan blocks) and instance edge detection branch (yellow blocks). Semantic edge branch generates semantic edges $f_S(I;\theta,\omega)$ by fusing multi-level features, and semantic edge loss $L_S$ is used to supervise Side5 output and the fused output. Object detection branch generates instance proposals $f_O(I;\theta,\varphi)$ through a region-based method and is supervised by object detection loss $L_O$. Based on prior information obtained from the above two branches, instance edge detection branch predicts instance edges $f_I(I;\theta,\gamma)$ for \emph{instances} and is supervised by instance edge loss $L_I$. (b) Illustration of instance edge detection branch. The compressed feature and selected semantic-level edge map which are obtained from semantic edge branch would be concatenated together. Then an instance proposal from $f_O(I;\theta,\varphi)$ is used to crop this concatenated feature. Finally, the cropped prior feature is fed into a transformation block to generate the instance-level edge map for this instance. Best viewed in color.}
\label{fig1-overall-architecture}
\end{figure*}

\section{Panoptic Edge Detection}

\subsection{Task Definition}
\label{section 3.1: Task Definition}

Given a category set $S = \{ S_{st} , S_{in}\}$ with $K$ categories, $S_{st}$ denotes the $K_{st}$ categories of \emph{stuff}, and $S_{in}$ represents the $K_{in}$ categories of \emph{instance} ($K=K_{st}+K_{in}$). \emph{Stuff} is the amorphous regions in a scene like sky, building or road, and \emph{instance} is the countable objects such as car, person. Panoptic edge detection (PED) task aims to outline all the labeled \emph{stuff} and \emph{instance}. Particularly, for a pixel $I_p$, the task aims to assign a correct label $Y(I_p)=\{Y_S(I_p), Y_I(I_p)\}$. Here, $Y_S(I_p)$ is the semantic label for \emph{stuff} categories. $Y_S(I_p)=(C_p^1, C_p^2,..., C_p^{K_{st}})$, where $C_p^k$ ($C_p^k \in [0,1]$) denotes the probability of edge.
For the instance label $Y_I(I_p)$, it is defined as $Y_I(I_p)=\{(Z_p^1, k_1),...,(Z_p^i, k_i),...,(Z_p^n, k_n)\}$, where $Z_p^i$ denotes the instance id, $k_i$ ($k_i \in S_{in}$) denotes the corresponding instance category. $Y_I(I_p)$ can be an empty set if the pixel is not associated with any instance.

\subsection{Panoptic Edge Network (PEN)}
Our ultimate optimization objective is to minimize the loss between ground truth $Y$ and prediction $f(I)$ for an image $I$. In our model, we disassemble panoptic edge detection task to semantic edge detection, object detection and instance edge detection. Specifically, we factorize the $f(I)$ into semantic edges $f_S(I)$, object bounding boxes $f_O(I)$ and instance edges $f_I(I)$. We devise a three-branch pipeline and take ResNet-50 \cite{he2016deep} as our backbone with some minor modifications as in~\cite{yu2017casenet} whose parameters are denoted as $\theta$. Figure \ref{fig1-overall-architecture}(a) shows the overall architecture of PEN model.

Semantic edge detection branch predicts semantic-level boundaries $f_S(I; \theta, \omega)$ by fusing multi-level side features as in DFF~\cite{hu2019dynamic} and its parameters are denoted as $\omega$. Object detection branch generates instance proposals $f_O(I; \theta, \varphi)$ through a region-based object detection method following Faster R-CNN~\cite{ren2015faster} and the parameters are represented by $\varphi$. Instance edge detection branch leverages semantic-level boundaries produced by semantic edge branch and instance bounding box proposals produced from object detection branch as prior information to generate instance-level boundaries $f_I(I; \theta, \gamma)$ for instance categories, and this branch is parameterized by $\gamma$. Similarly, we can disassemble our final optimization objective to three losses corresponding to the above three branches:
%
\begin{equation}
\mathcal{L} = \alpha_1\mathcal{L}_S+\alpha_2\mathcal{L}_O+\alpha_3\mathcal{L}_I
\label{eq.total loss L}
\end{equation}
where $\mathcal{L}_S, \mathcal{L}_O, \mathcal{L}_I$ denote the semantic edge loss, object detection loss and instance edge loss, respectively. $\alpha_1, \alpha_2, \alpha_3$ are the corresponding weighting factors used to balance three losses.

 \subsubsection{Semantic Edge Detection Branch}
 In this branch, we predict semantic-level boundaries for both stuff categories $S_{st}$ and instance categories $S_{in}$. Semantic results of \emph{instances} will later be used as prior knowledge for instance edge detection branch to predict instance-level boundaries.

 Taken an image $I$ as input, semantic branch learns a function $f_S(I;\theta,\omega)$ to map $I$ to semantic edge maps $f_S: I \to (C^1, C^2, ... , C^{K})$. PEN combines category-wise edge activations from high-level features with fine edge details from low-level features by fusing the multi-level side responses following DFF architecture, as shown in Figure~\ref{fig1-overall-architecture}(a). Specifically, low-level side features Side1$\sim$Side3 and high-level side output Side5 are concatenated using shared concatenation~\cite{yu2017casenet}. Then they are fused using a grouped $1 \times 1$ convolutional layer to produce semantic-level boundaries with $K$ channels. Semantic supervisions are imposed on Side5 output and the fused output.
We adopt the reweighted cross-entropy loss  to calculate multi-label loss for Side5 and the fused output. Given a semantic label $Y_S(I)$ and prediction $f_S(I;\theta, \omega)$, we have semantic edge loss $\mathcal{L}_{S}(I)$:
 \begin{equation}
 \mathcal{L}_{S}(I) = \sum\limits_{p\in I}\{-\eta \cdot(1-Y_S(I_p))\cdot \log(1- f_S(I_p;\theta, \omega)) - \bar{\eta} \cdot Y_S(I_p)\cdot \log(f_S(I_p;\theta, \omega)) \}
 \label{eq.L_S}
 \end{equation}
 where $\eta = |Y_S(I)_+|/ |Y_S(I)|$, $\bar{\eta} = |Y_S(I)_-|/ |Y_S(I)|$ are balance factors for edge, non-edge pixels.

 \subsubsection{Object Detection Branch}
 For object detection branch, bounding boxes of \emph{instance} are generated following the two-stage Faster R-CNN detector~\cite{ren2015faster}. Region Proposal Network (RPN) is utilized to generate region proposals which are then be used by Fast R-CNN detection sub-network to produce final object detection results. Then, given the object bounding box label $Y_{cls}$, $Y_{reg}$ and prediction $f_O(I;\theta, \varphi)$, we have object detection loss $\mathcal{L}_O(I)$ for object detection branch:
 %
 \begin{equation}
\mathcal{L}_O(I) = \mathcal{L}_{cls}(Y_{cls},f_O(I;\theta, \varphi_{cls})) +  \mathcal{L}_{reg}(Y_{reg},f_O(I;\theta, \varphi_{reg})).
\label{eq.L_O}
\end{equation}
The object detection branch, including RPN and Fast R-CNN, is extended from the fourth residual block (res4) of backbone (Figure \ref{fig1-overall-architecture}(a)) following~\cite{ren2015faster}.
PEN actually uses two res5 blocks: one res5 block used to extract features for Fast R-CNN in object detection branch and the other connected to Side5 to extract high-level features in semantic edge branch.
In this way, we alleviate the mutual influence between object-based representation learning and edge-based representation learning in a single backbone network.



 \subsubsection{Instance Edge Detection Branch}
 Intuitively, the computationally efficient way is to find the boundaries of an object within its bounding box. Assume we have obtained the semantic-level boundaries $f_S(I;\theta,\omega)$ with $K$ channels and the concatenated features with $4K$ channels from semantic edge branch as well as instance proposals $f_O(I;\theta,\varphi)$ from object detection branch, as shown in Figure \ref{fig1-overall-architecture}(a). Instance edge branch aims at instantiating edge prediction from the prior information.

 We extend the instance edge branch from the concatenated hierarchical side features in the semantic edge branch.
 Figure \ref{fig1-overall-architecture}(b) shows the details of instance edge detection branch. For each instance proposal in $f_O(I;\theta,\varphi)$, we extract an edge map from $f_S(I;\theta,\omega)$ according to its category.
 Then the concatenated feature with $4K$ channels is compressed to a $4$-channel feature map using a $1 \times 1$ conv layer, which is then concatenated with the selected semantic-level edge map. After that, the bounding box corresponding to the instance proposal is used to crop the concatenated feature. Finally, a transformation block is used to generate the instance-level boundary according to the cropped prior feature for this instance. The transformation block is composed of 3 convolution layers, and the first 2 conv layers are followed by a Batch Normalization (BN)~\cite{ioffe2015batch} layer and a ReLU layer, which is detailed in Section \ref{section 4.3: Ablation Experiments}.


 As aforementioned, instance edge branch can be represented as a parameterized function $f_I(I;\theta, \gamma)$ where $\gamma$ serves as the parameters of the compression and transformation block. Given the instance label $Y_I(I)$ and prediction $f_I(I;\theta, \gamma)$, we have instance edge loss $\mathcal{L}_{I}(I)$ expressed as
 %
 \begin{equation}
 \mathcal{L}_{I}(I) = \sum\limits_{p\in I}\{-\eta \cdot(1-Y_I(I_p))\cdot \log(1- f_I(I_p;\theta, \gamma)) - \bar{\eta} \cdot Y_I(I_p)\cdot \log(f_I(I_p;\theta, \gamma)) \}
 \label{eq.L_I}
 \end{equation}
 where $\eta = |Y_I(I)_+|/ |Y_I(I)|$, $\bar{\eta} = |Y_I(I)_-|/ |Y_I(I)|$ are balance factors for edge, non-edge pixels.

 Instance edge branch will maximize the expectation of transforming semantic visual understanding to instance edge awareness which are highly correlated. Being closely supervised by different task objectives, they may be  confused by each other.
As shown in Figure \ref{fig1-overall-architecture}(a), the selected semantic-level edge map receives distracted gradient signals from semantic edge supervision and instance edge supervision. The semantic edge supervision produces gradient signals for the learning of semantic-level edge boundaries, while the instance edge supervision produces instance-specific gradient signals that try to enhance the boundary response of the current specific instance and weaken other instances' boundary responses within the same category.
 To address this discrepancy, we \emph{detach} the  semantic-level edge map from the computational graph to block the gradient from instance edge supervision when back-propagating the instance edge loss.
 To investigate the influence of such  \emph{detaching} on quality of semantic and instance edges, we conduct an ablation experiment in Section \ref{section 4.3: Ablation Experiments}.

 \subsubsection{Inference}
 During inference, semantic edge branch outputs semantic edge maps with $K$ channels, including $K_{st}$ channels for \emph{stuff} categories and $K_{in}$ channels for \emph{instance} categories. Object detection branch produces instance proposals for \emph{instance} categories. In instance edge branch, for each instance proposal obtained from object detection branch, the predicted class label of this instance is used to select the corresponding channel from the semantic edge activation maps (before the sigmoid activation function). Then the selected semantic edge activation map is concatenated with the compressed supplementary feature. The instance proposal is used to crop the concatenated feature which is then forwarded to the transformation block. Finally, the instance edge branch would output the instance edge prediction for each instance.


\subsection{Panoptic Edge Metric}


We devise \emph{panoptic dual F-measure} ($F^2$) as evaluation of panoptic edge detection performance, which is able to measure edge prediction quality and instance recognition quality at the same time. $F^2$ is formally defined as
\begin{equation}
    F^2 = F_{edge} \times F_{object}
    \label{eq.F^2}
\end{equation}
where $F_{edge}$ denotes MF(ODS) for edge detection, and $F_{object}$ denotes $F_1$ score for object detection.

For each stuff category, $F_{edge}$ is computed at image level as in SED task and $F_{object}$ is equal to $1$. Then, we have $F^2$ for stuff categories:
\begin{equation}
    F_{st}^2 = F_{edge}
\end{equation}

For each instance category, there are two steps to calculate $F_{in}^2$: 1) coarse-to-fine instance matching and 2) $F_{in}^2$ computation given the matches.

For an instance category of an image, we have $n$ ground truth instances and $m$ predicted instances. Each ground truth instance must be matched with at most one predicted instance. Firstly, coarse matching is performed to select at most top $t$ predicted instances with bounding box intersection over union (IoU) greater than $0.5$ for each ground truth instance. This gives a coarse matching result where each ground truth instance will be matched with at most $t$ predictions. Ground truth instances with no matched prediction will be counted as false negatives (FN). Secondly, fine matching is performed to select one prediction instance with largest F-measure between ground truth instance edge map and predicted instance edge map for ground truth instances matched with more than one prediction in the process of coarse matching. After coarse-to-fine matching, each ground truth instance will be matched with no more than one prediction. We perform the coarse-to-fine matching strategy to accelerate the instance matching process and set $t=2$ in our implementation.

The matching result splits the predicted and ground truth instances into three sets: true positives (TP), false positives (FP) and false negatives (FN), representing matched pairs of instances, unmatched predicted instances and unmatched ground truth instances, respectively. (The readers may kindly refer to supplementary materials for more details.) Given the TP set, $F_{edge}$ is computed over instance-level predicted and ground truth pairs. Given all three sets, $F_{object}$ is calculated as
\begin{equation}
    F_{object} = \frac{TP}{TP + \frac{1}{2} FP + \frac{1}{2} FN}
\end{equation}
Then, $F_{in}^2$ can be computed according to Equation (\ref{eq.F^2}).

\section{Experiments}
\subsection{ Datasets}
We use two datasets with both semantic-level and instance-level annotations, and convert segmentation labels to boundary labels following the method used in CASENet \cite{yu2017casenet}.

\vspace{-2mm}
\paragraph{Cityscapes~\cite{cordts2016cityscapes}}
Cityscapes dataset contains $5,000$ images divided
into $2,975$ training, $500$ validation and $1,525$ test images, respectively. The training set and validation set are used for training and testing respectively, due to the unavailable ground truth of test set. There are $19$ categories in total, including $11$ \emph{stuff} categories and $8$ \emph{instance} categories.

\vspace{-2mm}
\paragraph{ADE20K~\cite{zhou2017scene}}
ADE20K dataset has around $25,000$ images divided into $20,000$ training, $2,000$ validation and $3,000$ test images, all densely annotated with an open-dictionary label set containing $50$ \emph{stuff} categories and $100$ \emph{instance} categories. Considering the severe class imbalance in converted edge annotations, we select $10$ \emph{stuff} and $10$ \emph{instance} classes that are more balanced from top $25$ pixel density classes. After screening, we get $14,417$ training images including $89,324$ instances and $1,435$ validation images including $9,838$ instances, used for training and testing respectively.

\subsection{Implementation Details}
\paragraph{Configuration}
During training, we use random mirroring and random scaling with short size sampled from $[480, 1280]$ on Cityscpaes and $\{300, 375, 450, 525, 600\}$ on ADE20K. We also augment each training sample with random cropping with crop size $640\times 640$ on Cityscpaes and $352\times 352$ on ADE20K. During testing, the images with original size are used for Cityscapes and  for ADE20K the images are padded to a minimum multiple of $8$ larger than original, since the stride of the highest layer is $8$.
We set the base learning rate to $0.01$ for both Cityscapes and ADE20K. The ``poly'' policy is used for learning rate decay and warmup strategy is used in first $500$ iterations. The batch size, maximum iterations, momentum, weight decay are set to $4$/$8$, $96000$/$25000$, $0.9$, $1e-4$ for Cityscapes and ADE20K respectively. We set $\alpha 1$, $\alpha 2$ and $\alpha 3$ in Equation (\ref{eq.total loss L}) as $8$/$5$, $1$ and $0.03$ for Cityscapes and ADE20K respectively. The proposed network is built on ResNet-50 \cite{he2016deep} pretrained on ImageNet \cite{deng2009imagenet}, and trained using the multi-task loss in Section \ref{section 3.1: Task Definition} and optimized by SGD using PyTorch \cite{paszke2017automatic}\footnote{Codes will be released soon.}. All experiments are performed using $4$ NVIDIA TITAN Xp($12$GB) GPUs.


\vspace{-2mm}
\paragraph{Baselines}
We design two competitive baselines. The first is denoted as \emph{CASENet+Faster}, which synthesizes semantic edge detection results from CASENet~\cite{yu2017casenet} and bounding box proposals from Faster R-CNN~\cite{ren2015faster}.  We train CASENet for both \emph{stuff} and \emph{instance} categories and Faster R-CNN for \emph{instance} categories separately, and then the bounding boxes generated by Faster R-CNN are used to crop the semantic edge maps of \emph{instance} categories to get the corresponding instance edge map.

The second is denoted as \emph{Panoptic FPN*}, which is adapted from Panoptic Feature Pyramid Network~\cite{kirillov2019panoptic} with some minor modifications. We replace the segmentation loss in semantic segmentation branch and mask branch with reweighted semantic edge loss and instance edge loss, respectively.


\begin{figure}
\begin{minipage}{0.56\linewidth}
\begin{center}
{\small
\makeatletter\def\@captype{table}\makeatother\caption{Ablation on training strategy and \emph{detaching} manner on Cityscapes. All $F$-scores are in \%.}
\label{ablation_stage_wise}
}
\end{center}
\end{minipage}
\hspace{3mm}
\begin{minipage}{0.4\linewidth}
\begin{center}
{\small
\makeatletter\def\@captype{table}\makeatother\caption{Ablation on structure of transformation block. All $F$-scores are in \%.}
\label{ablation_instance_branch}
}
\end{center}
\end{minipage}
\\
\begin{minipage}[t]{0.56\textwidth}
\begin{center}
{
  \begin{tabular}{ccc|ccc}
    \toprule
    \multirow{2.5}{*}{Models}   & \multirow{2.5}{*}{Detach}  & stuff    &\multicolumn{3}{c}{instance}
     \\
    \cmidrule(r){3-6}
    & & $F^2$ & $F_{e}$ & $F_{o}$ & $F^2$ \\
    \midrule
    Stage-wise & - &  76.1   &  67.6 &  54.9 &36.9    \\
    End-to-end &  N &  76.4  &  63.9 &  55.3 &35.8    \\
    End-to-end    & Y &   \textbf{76.5}   &  \textbf{67.8} &  \textbf{55.5} &\textbf{37.6}   \\
    \bottomrule
  \end{tabular}
}
\end{center}
\end{minipage}
\hspace{3mm}
\begin{minipage}[t]{0.4\textwidth}
\begin{center}
{
  \begin{tabular}{cccc}
    \toprule
    \multirow{2.5}{*}{Models}    &\multicolumn{3}{c}{instance}
     \\
    \cmidrule(r){2-4}
    &   $F_{e}$ & $F_{o}$ & $F^2$ \\
    \midrule
    1-conv   &  60.9 &  55.7 &34.0    \\
    2-conv   & 58.3 & \textbf{55.8} &   32.6\\
    3-conv    &  \textbf{67.6} &  54.9 &\textbf{36.9}   \\
    \bottomrule
  \end{tabular}
}
\end{center}
\end{minipage}
\vspace{-3mm}
\end{figure}

\subsection{Ablation Experiments}
\label{section 4.3: Ablation Experiments}
We first investigate the effect of \emph{detaching} and end-to-end joint training, and then compare different implementations of transformation block in instance edge branch. All ablation experiments use the same settings with ResNet-50 backbone and run on Cityscapes.

\paragraph{Effect of Detaching}
We first test the effect of detaching the semantic-level edge map from the computational graph on performance of semantic and instance edge detection, and show results in Row $2$ and $3$ in Table~\ref{ablation_stage_wise}. Whether detaching or not has inconspicuous effect on semantic edges for \emph{stuff} categories, but for \emph{instance} categories, detaching has a striking enhancement on edge quality $F_{edge}$ from $63.9\%$ to $67.8\%$. The results coincide with our expectation that blocking the gradient from instance edge supervision by detaching the semantic-level edge activation from the computational graph can sidestep the discrepancy of distracted gradient signals from semantic edge supervision and instance edge supervision, thus improves the edge detection performance for \emph{instance} categories.

\paragraph{Effect of End-to-End Joint Training}
We then compare performance of end-to-end joint training with stage-wise training for the proposed PEN model. As PEN can be disassembled to three branches and instance edge branch takes in prior information from semantic edge branch and object detection branch, we can separate the training procedure into two stages and train the whole network in a stage-wise way. In the first stage, we jointly train semantic edge branch and object detection branch; in the second we only fine-tune instance edge branch by keeping backbone and other two branches frozen. We can make following observations from Row $1$ and $3$ in Table~\ref{ablation_stage_wise}. For \emph{stuff} categories, end-to-end joint training improves semantic edge performance by $0.4\%$ over $F_{edge}$ or $F^2$. For \emph{instance} categories, end-to-end strategy consistently outperforms stage-wise training over edge quality $F_{edge}$, object recognition quality $F^2$ and overall metric $F^2$ by $0.2\%$, $0.6\%$ and $0.5\%$ respectively. The results indicate end-to-end joint training can improve the performance for each task.

\paragraph{Structure of Transformation Block}
We also compare different structures of transformation block: a neural network with $1$, $2$ and $3$ conv layers\footnote{In Table \ref{ablation_instance_branch}, \emph{1-conv} denotes a $1 \times 1$ conv layer with output channel equals $1$. \emph{2-conv} denotes a structure with two $1 \times 1$ conv layers with output channels equals $16$ and $1$, and the first one is followed by a BN and a ReLU layer. \emph{3-conv} denotes a sub-network with consecutive $1 \times 1$, $3 \times 3$ and $1 \times 1$ conv layers with output channels equal $16$, $16$ and $1$ respectively, and the first 2 conv layers are followed by a BN and a ReLU layer.}. We conduct the experiment using stage-wise training described in the previous paragraph. As shown in Table~\ref{ablation_instance_branch}, \emph{$3$-conv} gives the best $F^2$ on instance edge detection. Given this, we select the \emph{$3$-conv} implementation for all experiments.

\begin{table}[t]
  \caption{Comparison with baselines on Cityscapes. All $F^2$ scores are measured by \%.}
  \label{SOTA-table-city}
  \resizebox{\textwidth}{!}{
  \centering
  \begin{tabular}{c|c|c|c|c|c|c|c|c|c|c|c|c|c|c|c|c|c|c|c||c}
    \toprule
    Models     & road  & side. & buil. & wall & fence  &pole & light &sign & vege. &terrain &sky &person &rider &car &truck &bus &train &mot. &bike &mean \\
    \midrule \midrule
    CASENet+Faster &  88.2	&82.8	&86.2	&46.1	&46.3	&81.6	&73.6	&80.2	&90.1	&63.0 &89.9	&35.6	&30.9	&45.4	&6.5	&7.4	&0.8	&15.6	&28.1	&52.5
      \\
    Panoptic FPN*    &76.2	&75.9	&79.0	&39.3	&43.0	&71.0	&60.1	&65.0	&84.3	&56.6	 &86.3	&34.1	&36.6	&45.9	&20.1 &	30.8	&\textbf{23.4}&	25.1&	27.7	&51.6
       \\
    PEN(Ours)  &\textbf{89.8}	&\textbf{83.1}	&\textbf{86.6}	&\textbf{47.5}	&\textbf{50.1}	&\textbf{81.8}	&\textbf{76.4}	&\textbf{80.5}	&\textbf{90.5}	&\textbf{65.3} &\textbf{90.0}	&\textbf{43.0}	&\textbf{39.3}	&\textbf{49.6}	&\textbf{29.7}	&\textbf{43.2}	&18.8	&\textbf{37.2}	&\textbf{39.8}	&\textbf{60.1}
   \\
    \bottomrule
  \end{tabular}
  }
  \vspace{-3mm}
\end{table}

\begin{figure}
\begin{minipage}{0.47\linewidth}
\begin{center}
{\small
\makeatletter\def\@captype{table}\makeatother\caption{Comparison with baselines w.r.t. mean value of $F_{edge}$, $F_{object}$ and $F^2$ for \emph{stuff} and \emph{instance} categories respectively on Cityscapes. All $F_{edge}$, $F_{object}$ and $F^2$ scores are in \%.}
\label{SOTA-other-table-city}
}
\end{center}
\end{minipage}
\qquad 
\begin{minipage}{0.47\linewidth}
\begin{center}
{\small
\makeatletter\def\@captype{table}\makeatother\caption{Comparison with baselines w.r.t. mean value of $F_{edge}$, $F_{object}$ and $F^2$ for \emph{stuff} and \emph{instance} categories respectively on ADE20K. All $F_{edge}$, $F_{object}$ and $F^2$ scores are in \%.}
\label{SOTA-table-ade20k}
}
\end{center}
\end{minipage}
\\
\begin{minipage}{0.47\linewidth}
\begin{center}
{\small
\begin{tabular}{cc|ccc}
        \toprule
        \multirow{2.5}{*}{Models}   &     {stuff}    &\multicolumn{3}{c}{instance}
         \\
        \cmidrule(r){2-5}
        &  $F^2$ & $F_{e}$ & $F_{o}$ & $F^2$ \\
        \midrule
        CASENet+Faster&75.3  &  38.0 &  49.2 &21.3    \\
        Panoptic FPN*   &67.0 &  51.5 &  \textbf{58.8} &30.5       \\
        PEN(Ours)     & \textbf{76.5}  &  \textbf{67.8} &  55.5 &\textbf{37.6}   \\
        \bottomrule
   \end{tabular}
}
\end{center}
\end{minipage}
\qquad 
\begin{minipage}{0.47\linewidth}
\begin{center}
{\small
\begin{tabular}{cc|ccc}
        \toprule
        \multirow{2.5}{*}{Models}   &     {stuff}    &\multicolumn{3}{c}{instance}
         \\
        \cmidrule(r){2-5}
        &  $F^2$ & $F_{e}$ & $F_{o}$ & $F^2$ \\
        \midrule
        CASENet+Faster& \textbf{58.2} & 49.9  &  29.7 & 14.8  \\
        Panoptic FPN*   & 51.2 & 43.8  &  34.2 &     14.5  \\
        PEN(Ours)     & 56.6  &  \textbf{61.8} &  \textbf{34.5} &\textbf{21.4}   \\
        \bottomrule
   \end{tabular}
}
\end{center}
\end{minipage}
\vspace{-4mm}
\end{figure}

\subsection{Comparison with  State-of-the-Arts}
\label{section4.4: Comparison with  State-of-the-Arts}
We compare with two baselines, \emph{CASENet+Faster} and \emph{Panoptic FPN*}, on Cityscapes and ADE20K datasets. We report $F^2$ score on each individual category. In addition, to compare edge detection performance for \emph{stuff} and \emph{instance} categories separately, we also report the mean value of $F_{edge}$, $F_{object}$ and $F^2$ on them respectively.

\begin{figure*}[t]
\centering
\resizebox{\textwidth}{!}{
\includegraphics[]{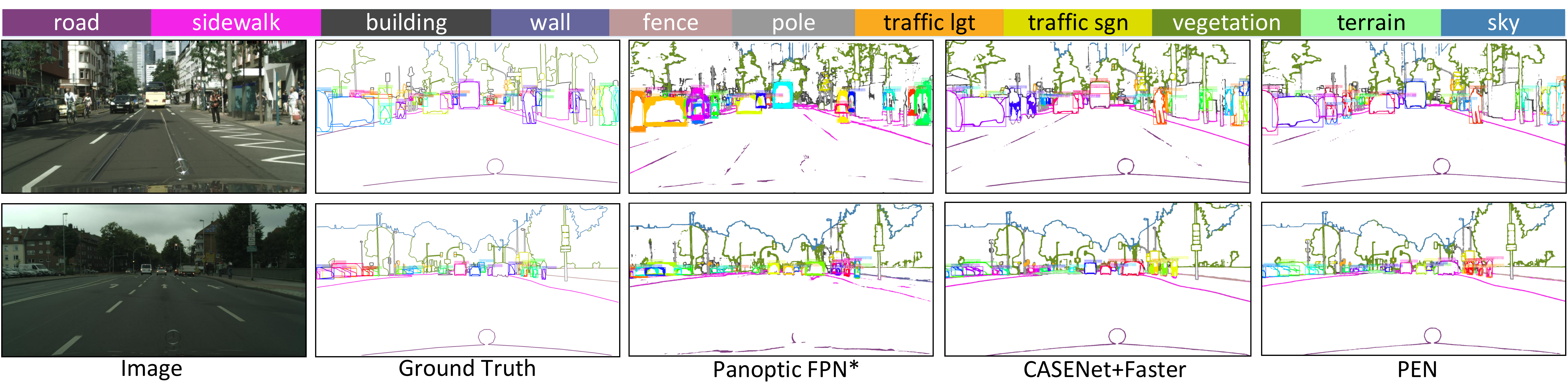}
}
\caption{Qualitative comparison on Cityscapes. From left to right:  ground truth, \emph{Panoptic FPN*}, \emph{CASENet+Faster} and PEN. Best viewed in color.}
\vspace{-4mm}
\label{fig2}
\end{figure*}

\subsubsection{Results on Cityscapes}
We first compare the proposed PEN model with \emph{CASENet+Faster} and \emph{Panoptic FPN*} on each individual category. Seen from Table \ref{SOTA-table-city}, PEN outperforms both  well established baselines and achieves $60.1\%$ on $F^2$. PEN is superior on almost all categories. Specifically, the $F^2$ of PEN is $7.5\%$ higher than \emph{CASENet+Faster} and $8.4\%$ higher than \emph{Panoptic FPN*} on average.

We then analyze performance for \emph{stuff} and \emph{instance} categories separately. As shown in Table \ref{SOTA-other-table-city}, compared with \emph{CASENet+Faster}, for \emph{stuff} categories, PEN achieves slightly higher performance ($1.2\%$) in $F_{edge}$ or $F^2$. However, for \emph{instance} categories, PEN outperforms \emph{CASENet+Faster} for a huge margin, $29.8\%$  in $F_{edge}$, $6.3\%$ in $F_{object}$ and $16.3\%$ in $F^2$ respectively. This indicates multi-task joint training can lead to better performance than training on semantic edge detection and object detection separately.

When comparing PEN with \emph{Panoptic FPN*}, PEN achieves $9.5\%$ and $16.3\%$ higher $F_{edge}$ for \emph{stuff} and \emph{instance} categories respectively. {\emph{Panoptic FPN*} performs unsatisfactorily possibly due to the lower resolution of hierarchical feature maps compared with the backbone PEN adopts and the \emph{ROI Align} operation adopted in mask branch of \emph{Panoptic FPN*}.} Seen from Figure \ref{fig2}, \emph{Panoptic FPN*} predicts very thick boundaries for instances, especially large objects, due to the \emph{resizing} in \emph{ROI Align}. Resizing objects to fixed size ($28 \times 28$) to predict boundaries and then resizing back to original would thicken the edges significantly. PEN evades this problem by abolishing all \emph{resizing} operations. However, \emph{Panoptic FPN*} outperforms PEN  $F_{object}$ by $3.3\%$ for \emph{instance} categories, mostly because of the adopted FPN architecture. Overall, PEN achieves $9.5\%$ and $7.1\%$ higher $F^2$ than \emph{Panoptic FPN*} for \emph{stuff} and \emph{instance} categories respectively, proving its superiority.

We visualize some prediction results in Figure \ref{fig2}. As we can see, \emph{Panoptic FPN*} gives unsatisfactory results for both semantic and instance edge detection, showing the inherent otherness between panoptic segmentation and panoptic edge detection. The panoptic segmentation oriented method does not work for panoptic edge detection. \emph{CASENet+Faster} is a more competitive baseline achieving almost the same good results with PEN in semantic edge detection. However, regarding instance edges, our PEN recognizes more instances and predicts clearer object boundaries. The readers may kindly refer to supplementary materials for more qualitative results.

\subsubsection{Results on ADE20K}
We compare the proposed PEN model with \emph{CASENet+Faster} and \emph{Panoptic FPN*} on \emph{stuff} and \emph{instance} categories separately in Table \ref{SOTA-table-ade20k}. For \emph{stuff} categories, PEN outperforms \emph{Panoptic FPN*} for a large margin and achieves almost the same good semantic edge results as competitive \emph{CASENet+Faster}. Regarding \emph{instance} categories, the proposed PEN model outperforms both baselines remarkably and achieves $21.4\%$ $F^2$, which well confirms its superiority on instance edge detection. The readers may kindly refer to supplementary materials for the $F^2$ score on each individual category and the qualitative results on ADE20K.

\section{Conclusion}
In this paper we introduced  a new panoptic edge detection (PED) task aiming at acquiring both semantic-level and instance-level object boundaries for \emph{stuff} and object \emph{instances}. We then proposed a multi-branch PEN framework which is jointly end-to-end trainable for solving PED task. To benchmark PED performance, we devised the panoptic dual F-measure ($F^2$) to evaluate quality of semantic and instance edges for \emph{stuff} and \emph{instances} in a uniform manner.

\bibliographystyle{unsrt}
\bibliography{neurips2019}

\begin{thebibliography}{10}

\bibitem{ren2015faster}
Shaoqing Ren, Kaiming He, Ross Girshick, and Jian Sun.
\newblock Faster r-cnn: Towards real-time object detection with region proposal
  networks.
\newblock In {\em Advances in neural information processing systems}, pages
  91--99, 2015.

\bibitem{deng2018learning}
Ruoxi Deng, Chunhua Shen, Shengjun Liu, Huibing Wang, and Xinru Liu.
\newblock Learning to predict crisp boundaries.
\newblock In {\em Proceedings of the European Conference on Computer Vision
  (ECCV)}, pages 562--578, 2018.

\bibitem{xie2015holistically}
Saining Xie and Zhuowen Tu.
\newblock Holistically-nested edge detection.
\newblock In {\em Proceedings of the IEEE international conference on computer
  vision}, pages 1395--1403, 2015.

\bibitem{liu2017richer}
Yun Liu, Ming-Ming Cheng, Xiaowei Hu, Kai Wang, and Xiang Bai.
\newblock Richer convolutional features for edge detection.
\newblock In {\em Proceedings of the IEEE conference on computer vision and
  pattern recognition}, pages 3000--3009, 2017.

\bibitem{liu2016learning}
Yu~Liu and Michael~S Lew.
\newblock Learning relaxed deep supervision for better edge detection.
\newblock In {\em Proceedings of the IEEE Conference on Computer Vision and
  Pattern Recognition}, pages 231--240, 2016.

\bibitem{prasad2006learning}
Mukta Prasad, Andrew Zisserman, Andrew Fitzgibbon, M~Pawan Kumar, and Philip~HS
  Torr.
\newblock Learning class-specific edges for object detection and segmentation.
\newblock In {\em Computer Vision, Graphics and Image Processing}, pages
  94--105. Springer, 2006.

\bibitem{yu2017casenet}
Zhiding Yu, Chen Feng, Ming-Yu Liu, and Srikumar Ramalingam.
\newblock Casenet: Deep category-aware semantic edge detection.
\newblock In {\em Proceedings of the IEEE Conference on Computer Vision and
  Pattern Recognition}, pages 5964--5973, 2017.

\bibitem{liu2018semantic}
Yun Liu, Ming-Ming Cheng, JiaWang Bian, Le~Zhang, Peng-Tao Jiang, and Yang Cao.
\newblock Semantic edge detection with diverse deep supervision.
\newblock {\em arXiv preprint arXiv:1804.02864}, 2018.

\bibitem{yu2018simultaneous}
Zhiding Yu, Weiyang Liu, Yang Zou, Chen Feng, Srikumar Ramalingam, BVK
  Vijaya~Kumar, and Jan Kautz.
\newblock Simultaneous edge alignment and learning.
\newblock In {\em Proceedings of the European Conference on Computer Vision
  (ECCV)}, pages 388--404, 2018.

\bibitem{hu2019dynamic}
Yuan Hu, Yunpeng Chen, Xiang Li, and Jiashi Feng.
\newblock Dynamic feature fusion for semantic edge detection.
\newblock {\em arXiv preprint arXiv:1902.09104}, 2019.

\bibitem{acuna2019devil}
David Acuna, Amlan Kar, and Sanja Fidler.
\newblock Devil is in the edges: Learning semantic boundaries from noisy
  annotations.
\newblock {\em arXiv preprint arXiv:1904.07934}, 2019.

\bibitem{kirillov2018panoptic}
Alexander Kirillov, Kaiming He, Ross Girshick, Carsten Rother, and Piotr
  Doll{\'a}r.
\newblock Panoptic segmentation.
\newblock {\em arXiv preprint arXiv:1801.00868}, 2018.

\bibitem{kirillov2019panoptic}
Alexander Kirillov, Ross Girshick, Kaiming He, and Piotr Doll{\'a}r.
\newblock Panoptic feature pyramid networks.
\newblock {\em arXiv preprint arXiv:1901.02446}, 2019.

\bibitem{he2017mask}
Kaiming He, Georgia Gkioxari, Piotr Doll{\'a}r, and Ross Girshick.
\newblock Mask r-cnn.
\newblock In {\em Proceedings of the IEEE international conference on computer
  vision}, pages 2961--2969, 2017.

\bibitem{he2016deep}
Kaiming He, Xiangyu Zhang, Shaoqing Ren, and Jian Sun.
\newblock Deep residual learning for image recognition.
\newblock In {\em Proceedings of the IEEE conference on computer vision and
  pattern recognition}, pages 770--778, 2016.

\bibitem{ioffe2015batch}
Sergey Ioffe and Christian Szegedy.
\newblock Batch normalization: Accelerating deep network training by reducing
  internal covariate shift.
\newblock {\em arXiv preprint arXiv:1502.03167}, 2015.

\bibitem{cordts2016cityscapes}
Marius Cordts, Mohamed Omran, Sebastian Ramos, Timo Rehfeld, Markus Enzweiler,
  Rodrigo Benenson, Uwe Franke, Stefan Roth, and Bernt Schiele.
\newblock The cityscapes dataset for semantic urban scene understanding.
\newblock In {\em Proceedings of the IEEE conference on computer vision and
  pattern recognition}, pages 3213--3223, 2016.

\bibitem{zhou2017scene}
Bolei Zhou, Hang Zhao, Xavier Puig, Sanja Fidler, Adela Barriuso, and Antonio
  Torralba.
\newblock Scene parsing through ade20k dataset.
\newblock In {\em Proceedings of the IEEE Conference on Computer Vision and
  Pattern Recognition}, pages 633--641, 2017.

\bibitem{deng2009imagenet}
Jia Deng, Wei Dong, Richard Socher, Li-Jia Li, Kai Li, and Li~Fei-Fei.
\newblock Imagenet: A large-scale hierarchical image database.
\newblock In {\em 2009 IEEE conference on computer vision and pattern
  recognition}, pages 248--255. Ieee, 2009.

\bibitem{paszke2017automatic}
Adam Paszke, Sam Gross, Soumith Chintala, Gregory Chanan, Edward Yang, Zachary
  DeVito, Zeming Lin, Alban Desmaison, Luca Antiga, and Adam Lerer.
\newblock Automatic differentiation in pytorch.
\newblock 2017.

\end{thebibliography}

\clearpage

\appendix
\section*{Appendix}

\section{Panoptic Edge Metric}

\begin{figure*}[h]
\centering
\resizebox{\textwidth}{!}{
\includegraphics[]{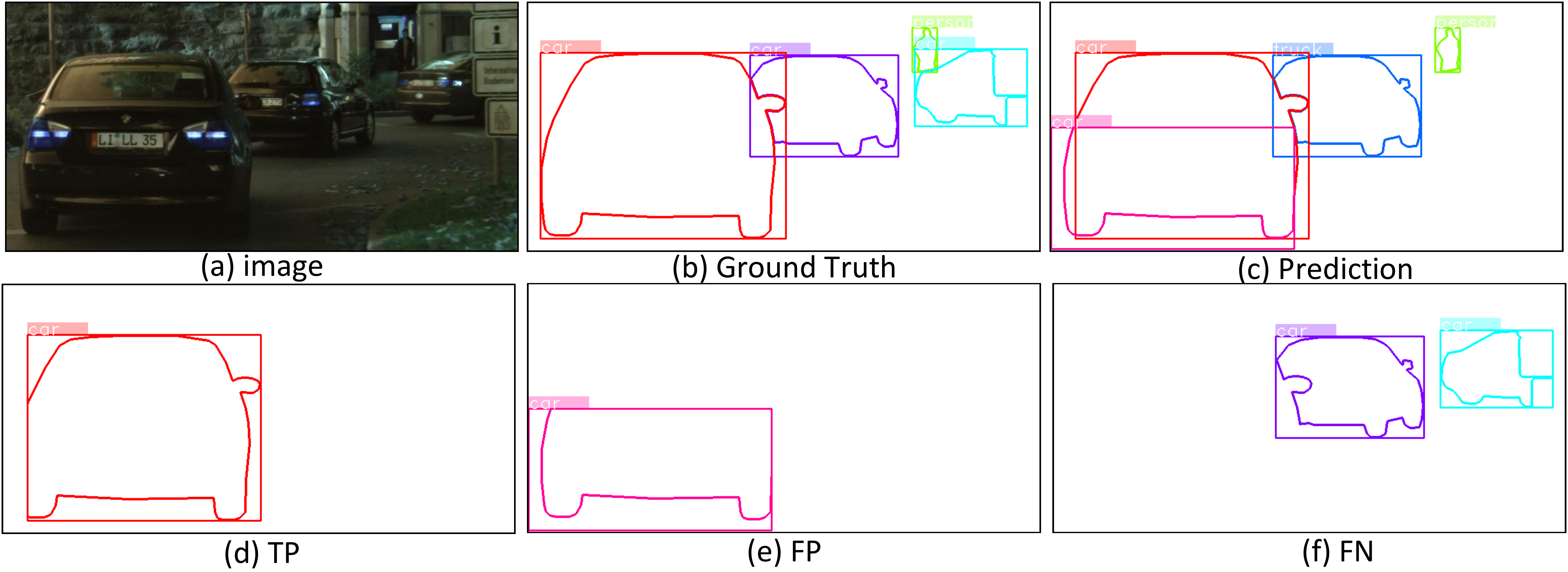}
}
\caption{Illustration of instance matching in $F^2$ computation. We take car as an example to show the matching results: true positive \emph{TP}, false positive \emph{FP}, and false negative \emph{FN}. Best viewed in color.}
\label{fig.evaluation metric}
\end{figure*}

To evaluate the instance matching results, we calculate $F^2$ for each category independently. Figure \ref{fig.evaluation metric} illustrates the instance matching procedure for the car category. Firstly, coarse matching is performed to select top $2$ predicted instances whose bounding boxes having  intersection over union (IoU) greater than $0.5$ for each ground truth car instance. Therefore, coarse matching gives $2$ matched predictions (two red cars in Figure \ref{fig.evaluation metric} (c)) for the red car in ground truth (Figure \ref{fig.evaluation metric} (b)), and another two unrecognized cars in Figure \ref{fig.evaluation metric} (b) fall in \emph{FN} set (Figure \ref{fig.evaluation metric} (f)). Then, in fine matching, F-measures are calculated between edge maps of the ground truth red car in Figure \ref{fig.evaluation metric} (a) and the two matched cars obtained from the coarse matching. The one with higher F-measure is selected as \emph{TP} and  the other is counted  in \emph{FP}, as shown in Figure \ref{fig.evaluation metric} (d) and (e).

\section{Additional Results on Cityscapes}

\begin{figure*}[!]
\centering
\resizebox{\textwidth}{!}{
\includegraphics[]{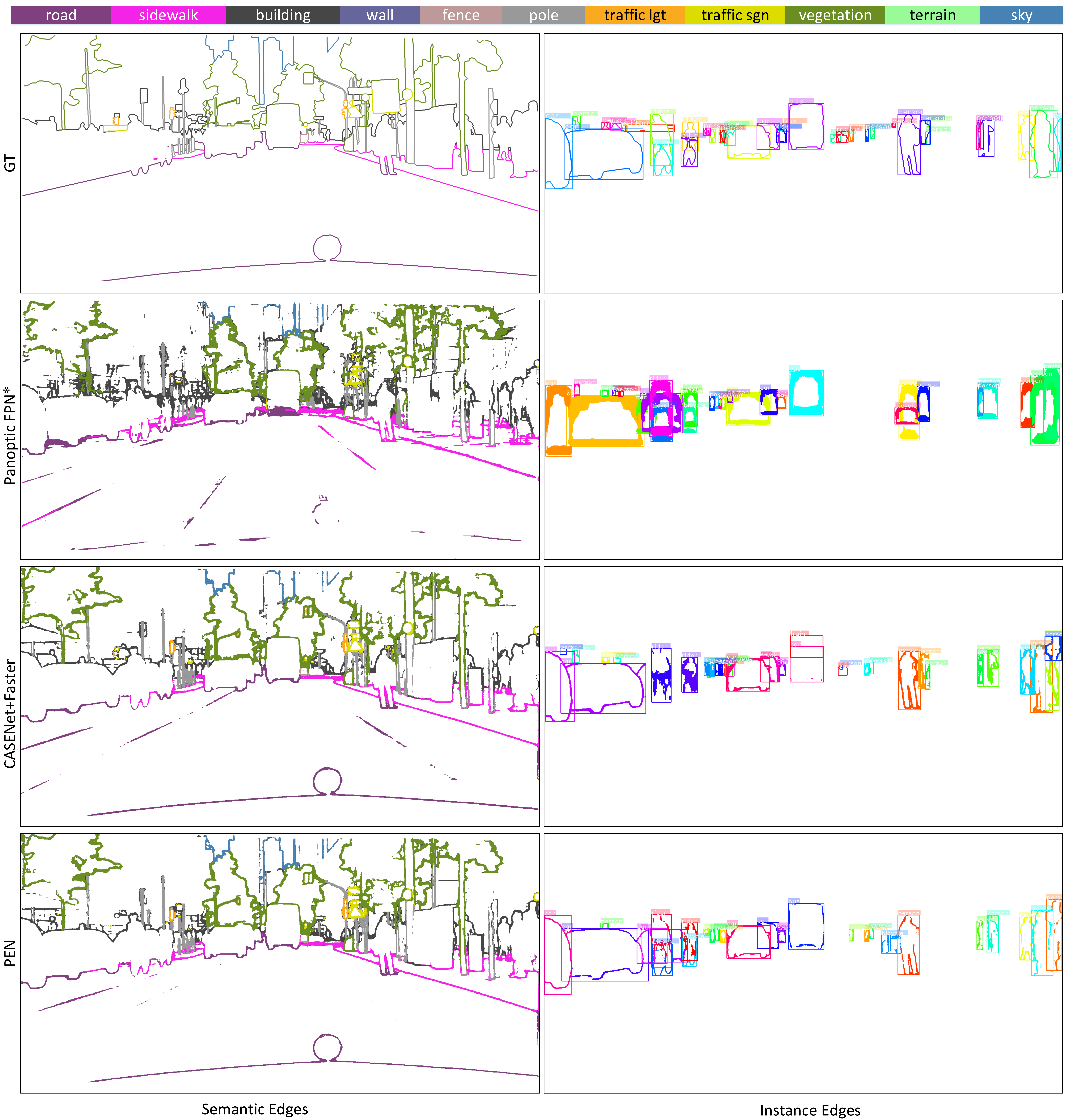}
}
\caption{Qualitative comparison of semantic and instance edges for \emph{stuff} (left column) and \emph{instance} (right column) respectively on Cityscapes. From top to bottom: ground truth, \emph{Panoptic FPN*}, \emph{CASENet+Faster} and PEN. Best viewed in color.}
\label{fig.visualization on Cityscapes}
\end{figure*}

In order to give  clearer comparison on the quality of semantic edges and instance edges detection for \emph{stuff} and \emph{instance} categories, we visualize the corresponding results  separately in Figure \ref{fig.visualization on Cityscapes}.

\section{Additional Results on ADE20K}

\subsection{Selected 20 Categories of ADE20K}

\begin{figure*}[h]
\centering
\resizebox{\textwidth}{!}{
\includegraphics[]{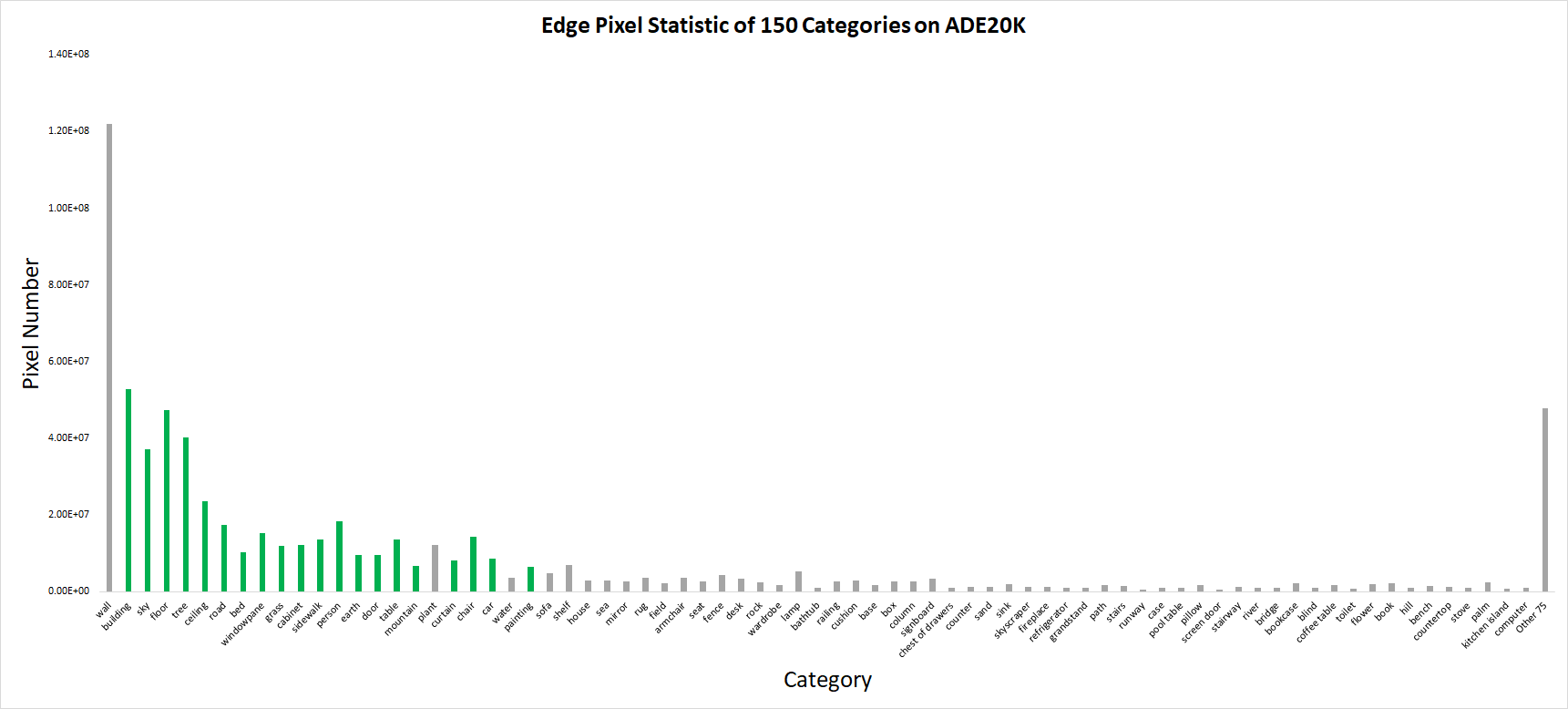}
}
\caption{Edge pixel statistics of the 150 categories on ADE20K with severe imblance.  We choose 20 categories with top  pixel numbers (the green ones).}
\label{fig.edge pixel statistics}
\end{figure*}

Figure \ref{fig.edge pixel statistics} shows edge pixel statistics of all 150 categories on ADE20K dataset. As can be observed, there is  severe class imbalance  for the annotated edge pixels. Specifically, the first category (wall) occupies  the large majority. From left to right, the number of edge pixels shows a decreasing trend  (the last number shows the  merged  statistics  of the rest 75 categories). Therefore, we choose $10$ \emph{stuff} and $10$ \emph{instance} classes that are more balanced from the  $25$ classes with top pixel densities. In particular, the \emph{Stuff} classes include building, sky, floor, tree, ceiling, road, grass, sidewalk, earth and mountain; and the \emph{instance} classes include bed, windowpane, cabinet, person, door, table, curtain, chair, car and painting.

\subsection{Comparison with State-of-the-Arts}

\begin{table}[h]
  \caption{Comparison with baselines on ADE20K in $F^2$ score (\%).}
  \label{SOTA-table-ade20k}
  \resizebox{\textwidth}{!}{
  \centering
  \begin{tabular}{c|c|c|c|c|c|c|c|c|c|c|c|c|c|c|c|c|c|c|c|c||c}
    \toprule
    Models     &build. &	sky	&floor&	tree
    &ceil.	&road	&grass	&side.	&earth	&moun.&	bed&	wind.&	cabinet&	person&	door&	table	&curtain	&chair	&car	&paint.
 &mean \\
    \midrule \midrule
    CASENet+Faster & \textbf{ 66.1}	&\textbf{78.4}	&\textbf{70.3}	&\textbf{65.9}	&65.2	&\textbf{61.4}	&60.5	&\textbf{56.7}  	&16.5	&\textbf{40.9}   &13.0	&16.4	&8.9	&17.9	&7.7	&14.7	&17.3	&15.6	&11.1	&\textbf{25.7} &36.5
      \\
    Panoptic FPN*    &60.0 	&61.1	&54.0	&56.9	&58.3	&47.7	&52.0	&54.8	&\textbf{27.9}	&38.9	 &14.1	&14.4	&10.5	&15.0	&8.6 &	9.1	&17.1&	13.3&	17.8	&24.8 &32.8
      \\
    PEN(Ours)  &62.0	&75.1	&68.6	&64.0	&\textbf{66.7}	&57.1	&\textbf{62.0}	&53.3	&17.2	&39.6 &\textbf{23.6}	&\textbf{18.7}	&\textbf{18.2}	&\textbf{23.6}	&\textbf{11.5}	&\textbf{18.5}	&\textbf{24.4}	&\textbf{21.2}	&\textbf{28.8}	&25.3 &\textbf{39.0}
   \\
    \bottomrule
  \end{tabular}
  }
\end{table}

As shown in Table \ref{SOTA-table-ade20k}, we compare the proposed PEN model with \emph{CASENet+Faster} and \emph{Panoptic FPN*} on each individual category. PEN outperforms both  well established baselines and achieves $39.0\%$ $F^2$. Specifically, the $F^2$ of PEN is $2.5\%$ higher than \emph{CASENet+Faster} and $6.2\%$ higher than \emph{Panoptic FPN*} on average, proving its superiority.

\subsection{Additional Qualitative Results}
\begin{figure*}[h]
\centering
\resizebox{\textwidth}{!}{
\includegraphics[]{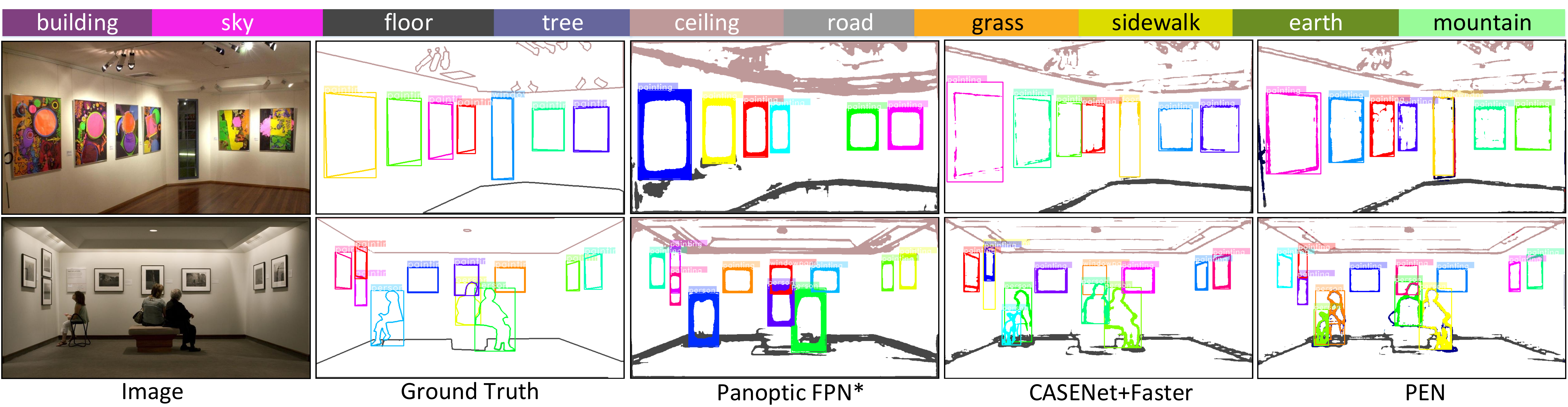}
}
\caption{Qualitative comparison on ADE20K. From left to right:  ground truth, \emph{Panoptic FPN*}, \emph{CASENet+Faster} and PEN. Best viewed in color.}
\label{fig.visualization on ade20k (panoptic)}
\end{figure*}

\begin{figure*}[!]
\centering
\resizebox{\textwidth}{!}{
\includegraphics[]{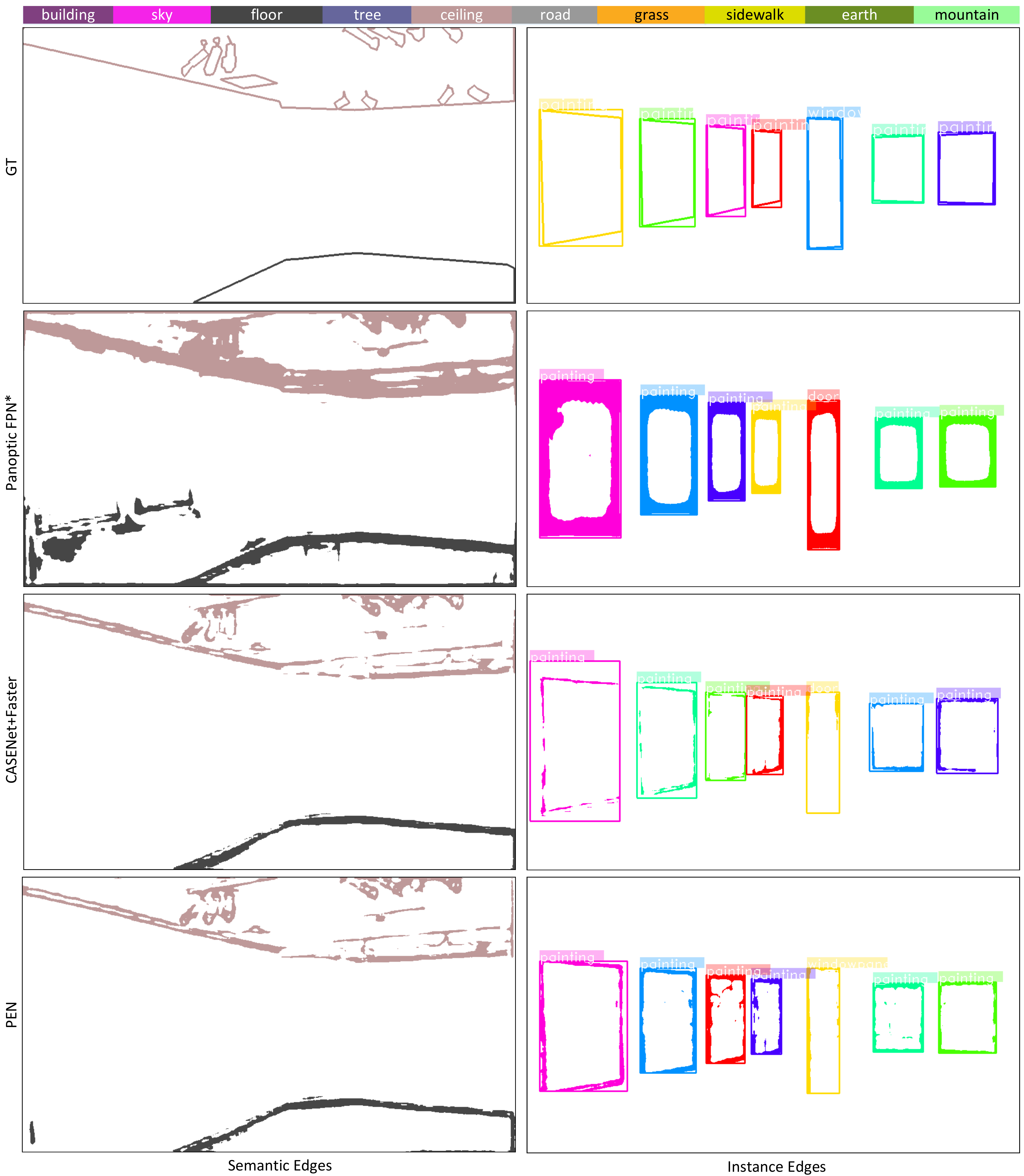}
}
\caption{Separate comparison of semantic and instance edges for \emph{stuff} (left column) and \emph{instance} (right column) respectively on ADE20K. From top to bottom: ground truth, \emph{Panoptic FPN*}, \emph{CASENet+Faster} and PEN. Best viewed in color.}
\label{fig.visualization on ade20k (semantic&instance)}
\end{figure*}

We visualize some PED   results in Figure \ref{fig.visualization on ade20k (panoptic)} and Figure \ref{fig.visualization on ade20k (semantic&instance)}. As one can see, \emph{Panoptic FPN*} gives unsatisfactory and over-thick edges for both semantic and instance edges. Compared with \emph{CASENet+Faster}, our PEN performs similarly well  on the semantic edges, but  much better  on the instance edges.

\end{document}